# Decoupling Multi-Contrast Super-Resolution: Self-Supervised Implicit Re-Representation for Unpaired Cross-Modal Synthesis

Yinzhe Wu, *Student Member, IEEE*, Hongyu Rui, Fanwen Wang, *Student Member*, Jiahao Huang, Zhenxuan Zhang, Haosen Zhang, Zi Wang, and Guang Yang, *Senior Member, IEEE*

***Abstract*—Multi-contrast super-resolution (MCSR) is crucial for enhancing MRI but current deep learning methods are limited. They typically require large, paired low- and high-resolution (LR/HR) training datasets, which are scarce, and are trained for fixed upsampling scales. While recent self-supervised methods remove the paired data requirement, they fail to leverage valuable population-level priors. In this work, we propose a novel, decoupled MCSR framework that resolves both limitations. We reformulate MCSR into two stages: (1) an unpaired cross-modal synthesis (uCMS) module, trained once on unpaired population data to learn a robust anatomical prior; and (2) a lightweight, patient-specific implicit re-representation (IrR) module. This IrR module is optimized in a self-supervised manner to fuse the population prior with the subject's own LR target data. This design uniquely fuses population-level knowledge with patient-specific fidelity without requiring paired target-domain LR/HR training data or paired cross-modal HR reference-target training data. Here, 'unpaired' refers to the population-level training setting; at subject-specific inference, as in standard MCSR, the method uses a matched HR reference image from another contrast together with the subject's LR target image. By building the IrR module on an implicit neural representation, our framework is also inherently scale-agnostic. Our method demonstrates superior quantitative performance on different datasets, with exceptional robustness at extreme scales (16×, 32×), a regime where competing methods fail. Our work presents a data-efficient, flexible, and computationally lightweight paradigm for MCSR, enabling high-fidelity, arbitrary-scale reconstruction without the need for paired population-level supervision.**

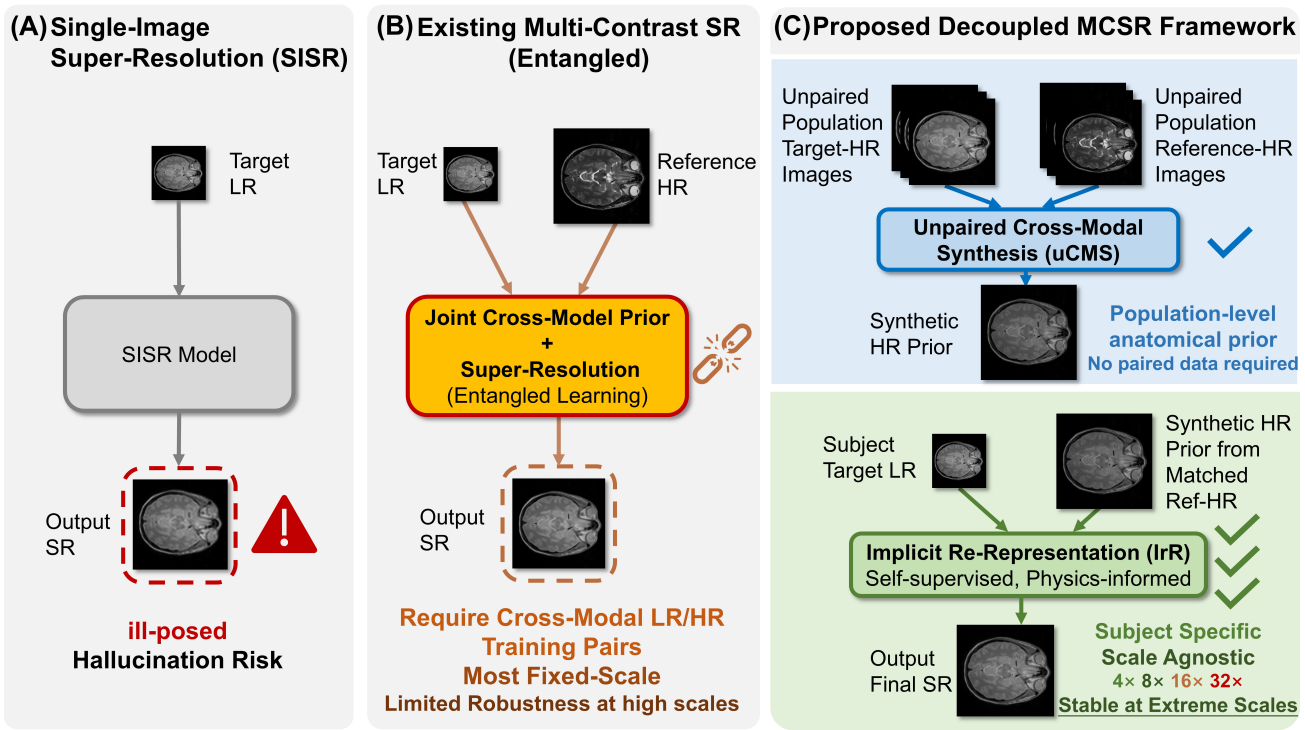

Fig. 1. **(A)** Single-image super-resolution (SISR) is ill-posed when only an LR target is available, leading to high hallucination risk due to insufficient constraints. **(B)** Existing multi-contrast super-resolution (MCSR) jointly learns cross-modal priors and SR, typically requiring paired cross-modal HR/LR training data, fixed scales, and showing limited robustness, particularly under severe information scarcity. **(C)** Proposed decoupled MCSR framework: population-level priors are learned via unpaired cross-modal synthesis (uCMS) from unpaired training data. At subject-specific inference, a synthetic target-domain HR prior generated from the matched HR reference image is fused with the subject's LR target image using implicit re-representation (IrR), enabling subject-specific, scale-agnostic, and data-consistent HR reconstruction through self-supervised, physics-informed optimization.

## I. INTRODUCTION

MAGNETIC Resonance Imaging (MRI) stands as a cornerstone of modern non-invasive clinical diagnostics, prized for its exceptional soft-tissue contrast and its capacity to provide rich anatomical and functional insights without the use of ionizing radiation. The diagnostic accuracy

This study was supported in part by Imperial College London President's PhD Scholarship and I-X. Zi Wang was supported in part by Imperial College London Seeds for Success Fund and Imperial College London Dame Julia Higgins Collaborative Research Fund. Guang Yang was supported in part by the ERC IMI (101005122), the H2020 (952172), the MRC (MC/PC/21013), the Royal Society (IEC\NSFC\211235), the NVIDIA Academic Hardware Grant Program, the SABER project supported by Boehringer Ingelheim Ltd, NIHR Imperial Biomedical Research Centre (RDA01), The Wellcome Leap Dynamic resilience program (co-funded by Temasek Trust), UKRI guarantee funding for Horizon Europe MSCA Postdoctoral Fellowships (EP/Z002206/1), UKRI MRC Research Grant, TFS Research Grants (MR/U506710/1), Swiss National Science Foundation (Grant No. 220785), and the UKRI Future Leaders Fellowship (MR/V023799/1, UKRI2738). (Yinzhe Wu and Hongyu Rui are co-first authors. Correspondence authors: Yinzhe Wu and Guang Yang)

Yinzhe Wu, Fanwen Wang, and Jiahao Huang are with the Department of Bioengineering and I-X, Imperial College London, London SW7 2AZ, United Kingdom and are also with the Cardiovascular Research Centre, Royal Brompton Hospital, London, United Kingdom. (email: yinzhe.wu18@imperial.ac.uk; fanwen.wang@imperial.ac.uk; j.huang21@imperial.ac.uk)

Hongyu Rui, Zhenxuan Zhang, Haosen Zhang and Zi Wang are with the Department of Bioengineering and I-X, Imperial College London, London SW7 2AZ. (email: hongyu.rui21@imperial.ac.uk; z.zhenxuan24@imperial.ac.uk; haosen.zhang24@imperial.ac.uk; zi.wang@imperial.ac.uk)

Guang Yang is with the Bioengineering Department and Imperial-X, Imperial College London, London W12 7SL, UK, and with the National Heart and Lung Institute, Imperial College London, London SW7 2AZ, UK, and with the Cardiovascular Research Centre, Royal Brompton Hospital, London SW3 6NP, UK, and with the School of Biomedical Engineering & Imaging Sciences, King's College London, London WC2R 2LS, UK. (email: g.yang@imperial.ac.uk)

of MRI is profoundly linked to its spatial resolution; the ability to visualize fine anatomical structures is paramount for the early detection and precise characterization of a wide spectrum of pathologies. However, acquiring high-resolution (HR) MRI is inherently time-consuming, which can cause patient discomfort and increase susceptibility to motion artefacts. This challenge is exacerbated by the fundamental trade-off between spatial resolution and signal-to-noise ratio (SNR). In many critical applications, particularly those with inherently low SNR (e.g., diffusion MRI) or those where signal averaging is impractical (e.g., functional MRI), a difficult compromise is required. Often, spatial resolution is intentionally sacrificed during acquisition to achieve a clinically acceptable SNR, leaving fine anatomical details unresolved.

In response to these intrinsic physical limitations, computational image super-resolution (SR) has emerged as a powerful and cost-effective post-processing strategy. Initial approaches focused on single-image super-resolution (SISR), where an HR image is inferred from a single low-resolution (LR) input. While SISR methods have demonstrated considerable success [1], the task remains a fundamentally ill-posed problem. In the context of medical imaging, this ill-posedness carries a significant clinical risk: SISR models, particularly generative ones, can introduce "hallucinated" structures [2], where anatomically plausible but factually incorrect details could lead to misdiagnosis. Moreover, enhanced image fidelity does not necessarily translate into more accurate downstream parametric maps [3], especially in quantitative imaging sequences that depend on multiple contrasts.

Modern clinical MRI protocols involve acquiring multiple images of the same anatomy using different sequences (e.g., T1-weighted (T1w), T2-weighted (T2w), Proton Density-weighted (PDw)), each highlighting distinct tissue properties while sharing the same underlying anatomy. This naturally presents a valuable opportunity for cross-modal enhancement, motivating the development of multi-contrast super-resolution (MCSR) techniques. The core principle of MCSR is to synergize this complementary information, typically by using an HR reference image from the reference contrast to guide the SR reconstruction of a target contrast [1]. By providing a high-fidelity anatomical template, the reference image constrains the SR process, reducing ambiguity and leading to a more faithful and reliable reconstruction than is possible with SISR [1], [4]. In this work, we target this standard MCSR setting, in which a subject-matched HR reference image from one contrast and an LR target image from another contrast are available at inference. Our contribution is not to remove this subject-specific multi-contrast input, which is intrinsic to MCSR, but to remove the need for paired population-level supervision during training. Specifically, the proposed framework does not require paired target-domain LR/HR samples or paired cross-modal HR reference-target samples in the training corpus.

Previous studies, particularly recent deep learning approaches, have demonstrated that leveraging complementary contrasts can significantly enhance SR performance [1], [4]. Zeng *et al.* proposed a two-stage architecture jointly addressing SISR and MCSR tasks [5], while Lyu *et al.* further introduced a joint feature space to better learn multi-contrast information [4]. More recently, various supervised methods – including multi-stage integration network [6], separation attentions [7], transformers [8], [9], [10], [11], diffusion models [12], conditional implicit network [13], [14], and convolutional dictionary model [15] – have further advanced the state of the art. In parallel, several unsupervised and self-supervised approaches have leveraged patient-specific, physics-informed learning via implicit neural representations (INR) [16], [17], [18], [19], [20].

Although many approaches have shown effectiveness, most existing MCSR methods assume fixed resolution settings [4], [11] and rely on large, perfectly paired LR/HR training datasets [1]. These assumptions rarely hold in clinical practice due to protocol variability and the scarcity of fully matched multi-modal datasets, which limits the robustness and generalizability of learned representations across subjects and modalities. Some studies have attempted to overcome the need for paired data by exploiting patient-specific features extracted from target LR and, when available, reference HR images of the same subject [16], [17]; however, such approaches prevent the model from learning population-level priors. Conversely, approaches that incorporate both patient-specific and population-level knowledge continue to rely on paired LR/HR datasets [18], thereby reintroducing the challenge of data scarcity.

In this work, we address this gap by introducing a framework that fuses population-level knowledge with patient-specific features, without requiring paired target-domain LR/HR training data or paired cross-modal HR reference-target training pairs.

From a modeling perspective, the challenge of MCSR fundamentally arises from the conflation of two distinct sub-problems: **(i)** learning a cross-modal anatomical prior that captures population-level structural regularities, and **(ii)** enforcing subject-specific fidelity and resolution recovery conditioned on sparse LR observations. Existing approaches typically entangle these two objectives within a single supervised or weakly supervised network, thereby inheriting both the data scarcity of paired training and the limited adaptability to subject-specific deviations.

Inspired by the separation of prior learning and inverse problem solving that underpins classical variational imaging [21], we propose to explicitly decouple MCSR into cross-modal synthesis (CMS) and target-domain re-representation. This reformulation allows population-level anatomical knowledge to be learned independently from large, unpaired datasets, while subject-specific super-resolution is performed via a self-supervised, physics-informed reconstruction process.

Specifically, we introduce a two-stage framework (Fig. 1) consisting of: **(1)** an unpaired cross-modal synthesis (uCMS) module trained once on population data to capture robust anatomical priors, and **(2)** a patient-specific implicit re-representation (IrR) module that enforces strict data consistency with the subject's acquired LR target image. Here, *"re-representation"* denotes that the target image is not represented from scratch, as in conventional INR methods [16], [17], [18], [19], [20], but is instead re-parameterized and re-optimized as a continuous implicit function conditioned on an existing synthesized HR estimate produced by the uCMS module. Importantly, the CMS output is treated not as a final reconstruction but as a soft prior that is explicitly corrected during re-representation, thereby mitigating the risk of hallucinated or modality-inconsistent structures. This design enforces a principled separation between population-level knowledge and patient-specific data fidelity, a concept central to variational imaging [21] but rarely made explicit in end-to-end deep learning MCSR [1]. Furthermore, by parameterizing the re-representation using an implicit neural representation conditioned on spatial coordinates rather than fixed grids, the proposed framework naturally supports arbitrary-scale SR.

**Our main contributions** are as follows:

**(1)** We introduce a top-down design that demonstrates MCSR can be modularly decomposed as the composition of CMS and target-domain re-representation, fully decoupling the training of these two components at different levels.

**(2)** We address the challenge of paired population-level supervision scarcity by fusing population-level priors with patient-specific information within a modular framework. The uCMS module is learned from unpaired population HR reference and HR target datasets, while the IrR module is optimized subject-specifically using only the acquired LR target image together with the subject's matched HR reference image. As a result, the framework does not require paired target-domain LR/HR training data or paired cross-modal HR reference-target training pairs.

**(3)** Our method supports scale-agnostic inference: by conditioning the INR on spatial coordinates instead of grid size, we achieve flexible, subject-specific super-resolution at any scale. Stress testing further demonstrated robust image fidelity even under extreme scale factors (16×, 32×).

While INRs and cross-modal priors have individually been explored for MRI super-resolution, existing approaches typically entangle population-level prior learning and patient-specific reconstruction within a single training pipeline, or rely on paired cross-modal or paired LR/HR supervision. In contrast, the proposed framework explicitly decouples population-level cross-modal synthesis from subject-specific super-resolution, allowing each component to be trained independently under fundamentally different data assumptions. This separation enables the incorporation of unpaired population priors without constraining patient-specific reconstruction to a fixed training distribution, distinguishing the proposed approach from prior INR-based MCSR and knowledge-transfer methods.

## II. Related Work

Multi-contrast super-resolution (MCSR) seeks to enhance LR images of a target modality by utilizing HR information from a reference modality. The core challenge lies in establishing effective cross-modal relationships to exploit their inherent similarities. Traditionally, this is achieved by extracting contrast-invariant features – such as image gradients [22], local covariance [23], and non-local similarity graphs [24] – which are often integrated into optimization-based approaches [22], [24] or interpolation filters [23].

### A. Supervised MCSR Approaches

Recent developments in deep learning-based approaches have further advanced the field. Zeng *et al*. proposed a two-stage architecture that jointly addresses SISR and MCSR [5]. Building on this foundation, Lyu *et al*. introduced a joint feature space to effectively learn complementary information across contrasts [4]. More recently, a series of methods have been developed to model cross-modal relationships, including multi-stage integration network [6], separation attentions [7], cross-attention transformers [8], [9], [10], [11], [25], [26], [27], diffusion models [12], [14], and convolutional dictionary model [15], all of which have further pushed the state of the art.

While these methods have progressively pushed the state-of-the-art, they all operate under the strong assumption that large, perfectly co-registered, and paired LR/HR training datasets are available. This reliance on paired data is a significant practical limitation, as such datasets are scarce and difficult to acquire.

A parallel line of inquiry has focused on addressing spatial misalignment using modules like displacement fields and deformable attention [11], [13], [15], [28], [29]. However, recent ablation studies report these modules offer *marginal fidelity gains* [15], [28], particularly at higher upscaling factors (e.g., 4×, 8×) [28]. Given that our work tends to demonstrate *in-scan* multi-modal enhancement at extreme scales (8× to 32×), the problem is dominated by the *massive information deficit* of upsampling, not minor spatial shifts. Therefore, in line with other state-of-the-art methods [14], we do not explicitly model spatial alignment.

### B. Implicit Neural Representations (INR) for MCSR

In parallel, several unsupervised and self-supervised approaches have emerged to overcome the need for paired data [16], [17], [18], [19], [20]. These methods typically leverage patient-specific, physics-informed learning via INRs. Instead of learning a mapping from a large population dataset, they optimize a network for each subject. A typical network learns a continuous mapping from spatial coordinates to signal intensities, guided by a self-supervised loss: the network's HR output, when downsampled, must match the subject's original LR input. To incorporate the cross-modal information, this implicit network is typically conditioned on features extracted from the HR reference image, often using an encoder module like a residual dense network.

A key advantage of employing an INR, shared by both these self-supervised approaches [16], [17], [18], [19], [20] and their supervised conditional counterparts [13], [14], is that the network is inherently scale-agnostic. Because the model is built upon spatial coordinates rather than a fixed-size grid, it can be queried at any arbitrary resolution during inference.

Focusing on the self-supervised methods [16], [17], [18], [19], [20], however, a critical limitation remains. While they successfully solve the paired-data problem, their exclusive reliance on patient-specific features prevents the model from learning powerful, population-level priors that can generalize across subjects and improve reconstruction robustness. In contrast to these conventional INR-based approaches, the proposed IrR module does not serve as the primary representation of the target image. Instead, it operates on top of a synthesized HR estimate generated by the uCMS module. The role of the implicit network is therefore to *re-represent* this existing population-level synthesis by correcting modality bias and enforcing subject-specific data fidelity, rather than constructing a representation solely from LR observations.

### C. Relating MCSR and Cross-Modal Synthesis (CMS)

A close conceptual relative of MCSR is cross-modal synthesis (CMS), which also aims to learn contrast-invariant features to generate one modality from another (e.g., T1w to T2w). The primary distinction is that MCSR is a super-resolution task that additionally uses the target domain's LR image as input. This LR input provides a critical constraint, grounding the reconstruction to be faithful to the subject-specific information, whereas CMS is purely a synthesis task.

Like MCSR, CMS has evolved significantly with deep learning. Many state-of-the-art methods are supervised, relying on large, perfectly-paired datasets. These often employ convolutional-GAN architectures like Pix2Pix [30], sometimes enhanced with features like edge-aware discriminators [31] or patch-based discriminators [32] to improve local detail synthesis. However, these supervised approaches are fundamentally limited by the same data-scarcity challenge that plagues paired MCSR.

To overcome this, unpaired CMS has become an extremely active area of research. These methods learn the translation without requiring co-registered image pairs. The most prominent example is CycleGAN [33], which enforces cycle-consistency and identity losses to learn the mapping. Other approaches, such as those using contrastive learning [34], have also shown success in extracting cross-modal features from unpaired data.

Recognizing this close relationship, some recent work has attempted to explicitly bridge CMS and MCSR. Feng *et al.* [28], for example, proposed a staged approach where they first train a *supervised* CMS model and then train a separate GAN-based module to refine the output using the target LR data. However, this method entangles the refinement module with the CMS output and, crucially, requires extensive paired target/reference HR data for its supervised training. This reintroduces the very data scarcity bottleneck that unpaired CMS methods were designed to solve, leaving a clear gap for a framework that can effectively fuse these tasks without relying on paired data.

## III. Methods

Our framework is founded on the principle that MCSR can be modularly decomposed into two distinct, independently optimized sub-problems: (1) learning a population-level prior of the target contrast from an unpaired reference, and (2) performing patient-specific, scale-agnostic super-resolution by fusing this prior with the subject's LR data.

This section first formally defines the MCSR problem in our unpaired setting. It then details the overall architecture of our proposed framework, followed by the descriptions of its two core components: the unpaired cross-modal synthesis (uCMS) module and the implicit re-representation (IrR) module.

### A. Problem Formulation

Let $X$ represent the domain of HR reference images (e.g., T2w) and $Y$ represent the domain of target-contrast images (e.g., PDw). We are given a set of unpaired images $I_x \in X$ and $I_y \in Y$. For a specific subject, we have access to a single HR reference image $I_{ref} \in X$ (at resolution $H \times W$) and a single LR target-contrast image $I_{lr} \in Y$ (at resolution $h \times w$), where $h \ll H$ and $w \ll W$. The ground-truth HR target image $I_{hr} \in Y$ (at $H \times W$) is unknown and unavailable for training.

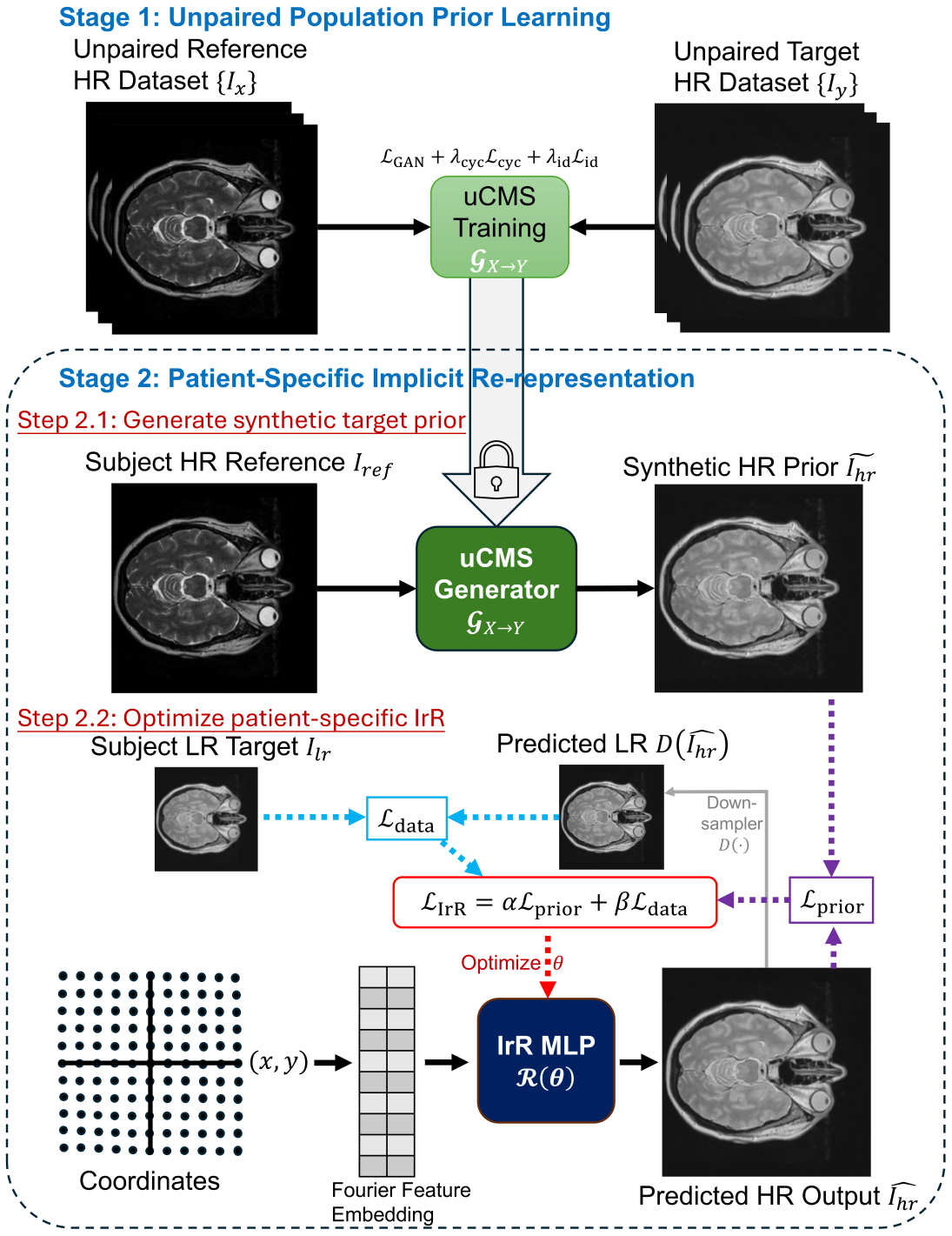

**Fig. 2.** Schematic of the proposed two-stage, decoupled MCSR framework. **Stage 1** (Unpaired Population Prior Learning): A uCMS generator ($\mathcal{G}_{X\to Y}$) is trained using a CycleGAN framework on unpaired population datasets ($\{I_x\}$ and $\{I_y\}$) to learn the cross-modal mapping. **Stage 2** (Patient-Specific Implicit Re-representation): For a new subject, the uCMS generator is frozen and used to create a synthetic HR prior ($\widetilde{I_{hr}}$) from the subject's HR reference ($I_{ref}$). A lightweight IrR MLP ($\mathcal{R}(\theta)$) is then optimized in a self-supervised manner. This optimization fuses the population-level prior (via $\mathcal{L}_{prior}$) with the subject's specific LR target ($I_{lr}$) (via $\mathcal{L}_{data}$), producing the final high-resolution output ($\widehat{I_{hr}}$). Operationally, Stage 2 is sequential at initialization: first generating the synthetic prior from the frozen uCMS module and then jointly constrained during IrR optimization by the prior-consistency and data-fidelity losses.

Our objective is to learn a function $\mathcal{F}$ that estimates the HR target image $\widehat{I_{hr}} \approx I_{hr}$, such that

$$\widehat{I_{hr}} = \mathcal{F}(I_{lr}, I_{ref}). \quad (1)$$

This must be achieved without access to any paired training data, meaning neither paired $(I_{lr}, I_{hr})$ volumes nor paired $(I_{ref}, I_{hr})$ volumes exist in our training corpus. As in standard MCSR, however, the subject-specific inference problem remains defined by a matched HR reference image $I_{ref}$ and an LR target image $I_{lr}$ for each test subject. Our framework solves this by factorizing $\mathcal{F}$ into two modules, $\mathcal{G}$ and $\mathcal{R}$, which are trained separately.

### B. Proposed Framework: Decoupled MCSR

At a high level, the proposed framework reformulates MCSR into two sequential but separately optimized components: (1) population-level cross-modal synthesis (uCMS), and (2) patient-specific target-domain re-representation (IrR). This decomposition first defines the overall decoupled framework conceptually, after which the two stages are described separately in terms of their roles, architectures, and optimization objectives. Importantly, the decoupling refers to how population-level prior learning is separated from subject-specific reconstruction optimization; it does not alter the standard MCSR inference setting, which still uses a matched HR reference image and LR target image for each subject.

It is important to emphasize that the uCMS module is not intended to directly produce the final super-resolved target image. As an unpaired synthesis model, uCMS may introduce residual artefacts or modality-inconsistent signals inherited from the reference contrast. In the proposed framework, uCMS serves exclusively as a population-level anatomical prior, whose output is subsequently corrected and constrained by the patient-specific IrR module through an explicit data fidelity term. This design ensures that the final reconstruction remains grounded in the subject's acquired measurements rather than the synthesized prior alone.

***1) Stage 1: Population Prior Learning (uCMS).***

We first train a generator $\mathcal{G}: X \to Y$ on the entire unpaired dataset $\{I_x\}$ and $\{I_y\}$. This module, implemented as a CycleGAN [33], learns the population-level mapping of anatomical features from the reference contrast (e.g., T2w) to the target contrast (e.g., PDw). Once trained, $\mathcal{G}$ is frozen during subject-specific fitting, so that population-level prior learning remains separated from patient-specific reconstruction optimization.

***2) Stage 2: Patient-Specific SR (IrR).***

For a new, individual subject, we generate a synthetic HR prior $\widetilde{I_{hr}} = \mathcal{G}(I_{ref})$. We then train a small, patient-specific implicit neural network $\mathcal{R}$ in a self-supervised manner. This module $\mathcal{R}$ learns to reconstruct the final $\widehat{I_{hr}}$ by fusing information from two sources: (1) the high-frequency anatomical structure from the "pseudo-HR" prior $\widetilde{I_{hr}}$ and (2) the subject-specific fidelity from the actual LR image $I_{lr}$.

The final reconstruction is the output of the optimized IrR module $\mathcal{R}$. An overview of this decoupled formulation is shown in Fig. 1, while the detailed architecture and optimization workflow are illustrated in Fig. 2.

## C. Module 1: Unpaired Cross-Modal Synthesis (uCMS)

To learn a population-level cross-modal anatomical prior without requiring paired data, we adopt an unpaired image-to-image translation strategy for the uCMS module. In this work, CycleGAN [33] is used as a representative and well-established unpaired synthesis model. While more recent unpaired synthesis techniques exist [34], CycleGAN [33] provides a controlled and interpretable baseline that allows us to isolate and study the effect of explicitly decoupling population-level prior learning from patient-specific reconstruction. Importantly, in the proposed framework, the uCMS output is not treated as a final reconstruction, but rather as a soft prior that is subsequently corrected by an explicit data fidelity constraint during implicit re-representation. This design makes the framework robust to moderate synthesis bias in the uCMS output and allows alternative unpaired CMS models to be readily substituted without altering the overall formulation.

***1) Architecture***

We implement the uCMS generator $\mathcal{G}$ (specifically $\mathcal{G}_{\mathcal{X}\to\mathcal{Y}}$) using a CycleGAN framework. **Generator ($\mathcal{G}$):** The generator adopts a U-Net architecture with residual connections. The encoder comprises four residual blocks (output channels: 32, 64, 128, 256) followed by a 512-channel bottleneck. Each residual block uses two `Conv2d-GroupNorm-Mish` units. Downsampling is achieved via $2 \times 2$ max pooling. The decoder mirrors this structure, using transposed convolutions for upsampling and incorporating skip connections. A final $1 \times 1$ convolution with a sigmoid activation produces the synthesized image. **Discriminator ($\mathcal{D}$):** We use a PatchGAN-style CNN with four $4 \times 4$ convolutional layers (stride 2). All layers except the first use batch normalization and LeakyReLU (slope 0.2). A global average pooling (GAP) layer and a fully connected sigmoid classifier predict patch-wise reality.

***2) Loss Function***

The uCMS module is trained using a combination of adversarial, cycle-consistency, and identity losses on the unpaired population data, as originally defined in [33].

$$\mathcal{L}_{uCMS} = \mathcal{L}_{GAN}(\mathcal{G}_{X\to Y}, \mathcal{D}_Y) + \mathcal{L}_{GAN}(\mathcal{G}_{Y\to X}, \mathcal{D}_X) + \lambda_{cyc}\mathcal{L}_{cyc} + \lambda_{id}\mathcal{L}_{id}, \quad (2)$$

where $\mathcal{L}_{GAN}$ is the standard adversarial loss (e.g., MSE loss), $\lambda_{cyc}$ and $\lambda_{id}$ are weighting hyperparameters. The cycle-consistency loss $\mathcal{L}_{cyc}$ ensures that $\mathcal{G}_{Y\to X}\mathcal{G}_{X\to Y}(I_x) \approx I_x$, and the identity loss $\mathcal{L}_{id}$ regularizes the generators to be identity mappings for their target domains (e.g., $\mathcal{G}_{X\to Y}(I_y) \approx I_y$).

## D. Module 2: Implicit Re-representation (IrR)

After the uCMS module $\mathcal{G}$ is frozen, we perform patient-specific super-resolution. The IrR module $\mathcal{R}$ is a standard coordinate-based INR parameterization, implemented as a multi-layer perceptron (MLP), but differs from a conventional INR in its role: rather than representing the target image from scratch from LR observations alone, it re-represents a synthesized target-domain prior under a joint prior-consistency and data-fidelity objective. Concretely, $\mathcal{R}$ learns a continuous mapping $\mathcal{R}: \mathbb{R}^2 \to \mathbb{R}$ from a 2D spatial coordinate $v = (x, y)$ to an intensity value $I(v)$.

***1) Architecture***

The IrR module $\mathcal{R}$ is intentionally lightweight, as it is optimized for each new subject.

**1. Fourier Feature Encoding:** To enable the MLP to learn high-frequency details, we first map the input coordinates $v$ using a Fourier feature mapping $\gamma(\cdot)$

$$\gamma(v) = [\cos(2\pi Bv), \sin(2\pi Bv)]^{\mathrm{T}}, \quad (3)$$

where $B$ is a matrix of frequencies sampled from a Gaussian distribution $\mathcal{N}(0, \sigma^2)$. The encoded coordinate $\gamma(v)$ serves as the input to the MLP.

**2. MLP Backbone:** The core of $\mathcal{R}$ is an 8-layer MLP with 256 hidden units per layer and Mish activations. The network $\mathcal{R}(v; \theta)$ with parameters $\theta$ outputs the predicted intensity $\widehat{I_{hr}}(v)$ at the continuous coordinate $v$.

***2) Self-Supervised Loss Function***

For a given subject, we optimize the weights $\theta$ of the IrR network $\mathcal{R}$ by minimizing a weighted loss that enforces consistency with both the population prior and the LR data. The total loss $\mathcal{L}_{IrR}$ is

$$\mathcal{L}_{IrR} = \alpha \cdot \mathcal{L}_{prior} + \beta \cdot \mathcal{L}_{data}, \quad (4)$$

where $\alpha$ and $\beta$ are scalar weights. We fix $\beta = 1.0$ throughout to keep the data-fidelity term as the reference scale of the objective, and tune only $\alpha$, since the trade-off is governed by the relative weighting between the prior-consistency and data-fidelity terms. $\alpha$ is selected on a held-out validation setting for each upsampling factor and then fixed across subjects (Section IV.E).

The two loss components are:

**1. Prior Consistency Loss** $(\mathcal{L}_{prior})$: This loss enforces that the reconstructed image $\widehat{I_{hr}}$ (generated by querying $\mathcal{R}$ at all HR coordinates $V_{hr}$) should resemble the synthetic prior $\widetilde{I_{hr}} = \mathcal{G}(I_{ref})$.

$$\mathcal{L}_{prior} = \sum_{v \in V_{hr}} \left\| \mathcal{R}(v; \theta) - \widetilde{I_{hr}}(v) \right\|_2^2 \quad (5)$$

This loss injects the rich anatomical information and population-level knowledge from the uCMS module. A direct pixel-wise loss to the raw HR reference image would not be meaningful here, because the reconstruction and the reference belong to different contrast domains; for this reason, the prior term is defined in the target domain through the synthesized target-domain prior $\tilde{I}_{hr} = \mathcal{G}(I_{ref})$, rather than through the raw reference image itself.

**2. Data Fidelity Loss** $(\mathcal{L}_{data})$: This self-supervised loss ensures the reconstruction is faithful to the subject's actual acquired data. It enforces that the network's output $\widehat{I_{hr}}$, when downsampled by a known operator $D(\cdot)$ (here, k-space truncation), must match the known LR input $I_{lr}$.

$$\mathcal{L}_{data} = \left\| D\big(\mathcal{R}(\cdot; \theta)\big) - I_{lr} \right\|_2^2 \quad (6)$$

This loss corrects for any subject-specific deviations or synthesis errors from the $\widetilde{I_{hr}}$ prior, grounding the final output in the patient's true data.

By optimizing $\mathcal{R}$ with this composite loss, the synthesized image from uCMS is re-represented as a continuous, coordinate-conditioned function that deviates from the initial population-level estimate only where required by the subject's acquired LR measurements. This distinguishes IrR from standard INR formulations, where the implicit network defines the image representation *ab initio*. As $\mathcal{R}$ is a coordinate-based function, inference can be performed at any arbitrary scale factor by querying the network on the desired coordinate grid.

## E. Training and Implementation

**uCMS Training:** A separate CycleGAN/uCMS model is trained for each dataset and its corresponding reference-to-target contrast pair; the synthesis network is not shared across datasets. Each model is trained for 50 epochs using the Adam optimizer with a batch size of 8. The generator learning rate is $1 \times 10^{-5}$ and the discriminator learning rate is $1 \times 10^{-6}$, with $\beta_1 = 0.5, \beta_2 = 0.999$. The training time for CycleGAN is around 16.7 sec per 2D slice.

**IrR Training:** For each subject, the IrR module is trained from scratch. We use 256-dimensional Fourier feature embeddings. The 8-layer MLP is optimized using Adam with a learning rate of $2 \times 10^{-4}$ and a cosine annealing scheduler. We use a batch size of 5000 randomly sampled coordinates per iteration. All coordinate inputs are scaled to $[-1, 1]$. Unless otherwise stated, the data-fidelity weight is fixed to $\beta = 1.0$ in all experiments. The prior-consistency weight $\alpha$ is selected on a held-out validation setting for each upsampling factor and then fixed across subjects, as further explained in Section IV.E.

The IrR module is optimized per test subject (patient-specific). Notably, the vanilla INR baseline (vanINR) uses the same coordinate-based neural representation architecture as the proposed IrR module for fair comparison.

Both modules were implemented in PyTorch and on a single NVIDIA GeForce RTX 3090 GPU (24 GB). On a single NVIDIA RTX 3090 (24 GB), the optimization time is around 55.7 sec per 2D slice under our default settings, with two IrR instances trained in parallel.

## F. Experimental Details

### 1) Datasets

Experiments were performed on three diverse, publicly-available multi-modal MRI datasets:

**IXI Dataset** [35]: This dataset contains multi-contrast brain MRI scans from healthy subjects collected from three different centers, each using different scan protocols. For our task, we used T2w images as the reference modality and PDw images as the target modality. We obtained 404 subjects for training and 174 subjects for testing.

**FastMRI Dataset** [36], [37]: This clinical patient cohort dataset provides multi-contrast knee MRI, and includes pathology-bearing cases. We used the fat-suppressed PDw (PD-FS) scans as the reference modality and the standard PDw scans as the target modality. We obtained 329 subjects for training and 141 subjects for testing.

**HCP Dataset** [38]**:** This dataset contains multi-contrast brain MRI scans from healthy subjects. For our task, we used T1 images as the reference modality and T2w images as the target modality. We obtained 127 subjects for training and 55 subjects for testing.

### 2) Data Preparation

To simulate the LR inputs $I_{lr}$ for our experiments, we adopted a degradation model based on k-space truncation, a standard practice in current MRI super-resolution studies [11], [15], [28].

For a given ground-truth HR image $I_{hr}$ (e.g., at $H \times W$ resolution), we first transformed it into the frequency domain via a 2D Fast Fourier Transform (FFT). We then simulated downsampling by applying a rectangular mask to perform a k-space center crop. This mask retains only the central $1/s$ fraction of frequencies in each dimension, where $s$ is the desired scale factor (e.g., 4×, 8×, 16×, or 32×), and zero-pads the rest. An inverse FFT was then applied to this truncated, low-frequency k-space data to produce the final LR image $I_{lr}$ (at $H \times W$).

This $I_{lr}$ image serves as the fidelity constraint for our IrR module. The downsampling operator $D(\cdot)$ described in our Methods (Sec III.D.2.2) is the precise analytic operator that performs this combined process of FFT, k-space center cropping, and inverse FFT.

*3) Evaluation:* We evaluate the quality of the super-resolved images $(\widehat{I_{hr}})$ against the ground-truth HR images $(I_{hr})$ using three common image fidelity metrics: (a) Peak Signal-to-Noise Ratio (PSNR), (b) Structural Similarity Index Measure (SSIM), and (c) Learned Perceptual Image Patch Similarity (LPIPS)

# IV. Results

We conducted a comprehensive qualitative and quantitative evaluation of our proposed framework against several state-of-the-art (SOTA) methods. These include supervised MCSR methods ((1) dictionary-based method A2CDic [15], (2) Transformer-based method McMRSR [8] and DANCE [11], (3) diffusion-model-based method DiffMSR [12], and (4) spatial alignment network based method SAN [29]), as well as an SISR SOTA method (SwinIR [39]). In addition, we evaluated the proposed model against the vanilla variants of its two constituent parts: an unpaired CMS model (CycleGAN) [33] and a self-supervised INR model (vanINR).

TABLE I
QUANTITATIVE COMPARISON OF OUR PROPOSED METHOD ("OURS") AGAINST STATE-OF-THE-ART METHODS AT THE SCALE FACTORS OF 4× AND 8×. PERFORMANCE IS EVALUATED ON IXI, HCP AND FASTMRI USING PSNR ↑, SSIM ↑, AND LPIPS ↓ BY THEIR MEAN AND (STANDARD DEVIATION). THE BEST RESULT FOR EACH METRIC IS HIGHLIGHTED IN RED. THE SECOND BEST IS HIGHLIGHTED IN BLUE.
* DENOTES STATISTICALLY SIGNIFICANT DIFFERENCE FROM OURS FOR THE SAME DATASET, SCALE FACTOR AND METRIC (P<0.01).

| | | | IXI | | | HCP | | | FastMRI | | |
|---|---|---|---|---|---|---|---|---|---|---|---|
| | | | PSNR ↑ | SSIM ↑ | LPIPS ↓ | PSNR ↑ | SSIM ↑ | LPIPS ↓ | PSNR ↑ | SSIM ↑ | LPIPS ↓ |
| 4× | MCSR | Ours | 35.16 (2.08) | .9319 (.0133) | .0123 (.0052) | 30.75 (0.88) | .9070 (.0151) | .0232 (.0066) | 27.97 (2.21) | .6708 (.0997) | .3841 (.0703) |
| | | A2CDic [15] 2025 | 33.39* (2.06) | .9451* (.0180) | .0525* (.0173) | 30.08* (0.80) | .9085* (.0110) | .0549* (.0089) | 26.20* (2.53) | .6026* (.1096) | .4301* (.0988) |
| | | DANCE [11] 2025 | 32.09* (2.05) | .9298* (.0218) | .0384* (.0111) | 29.54* (0.76) | .9022* (.0108) | .0399* (.0078) | 25.66* (2.48) | .5848* (.1088) | .3781* (.0850) |
| | | DiffMSR [12] 2024 | 31.58* (1.72) | .9248* (.0199) | .0780* (.0220) | 30.03* (0.84) | .9058* (.0116) | .0767* (.0125) | 26.14* (2.48) | .5971* (.1089) | .4350* (.0913) |
| | | McMRSR [8] 2022 | 33.21* (2.00) | .9415* (.0185) | .0591* (.0201) | 29.81* (0.86) | .9002* (.0192) | .0657* (.0105) | 26.27* (2.52) | .6021* (.1095) | .4330* (.0963) |
| | | SAN [29] 2022 | 32.50* (1.87) | .9319* (.0193) | .0725* (.0212) | 29.75* (0.80) | .8984* (.0108) | .0904* (.0133) | 26.00* (2.56) | .5930* (.1057) | .4399* (.0945) |
| | SISR | SwinIR [39] 2021 | 29.96* (1.42) | .8717* (.0201) | .1265* (.0277) | 28.03* (0.68) | .8481* (.0124) | .1563* (.0141) | 25.68* (2.32) | .5767* (.1068) | .4590* (.0838) |
| 8× | MCSR | Ours | 32.53 (2.47) | .8842 (.0194) | .0160 (.0068) | 28.13 (0.64) | .8576 (.0204) | .0196 (.0036) | 24.35 (1.77) | .5493 (.0991) | .5773 (.0707) |
| | | A2CDic [15] 2025 | 29.03* (1.72) | .8653* (.0317) | .1056* (.0283) | 27.69* (0.87) | .8463* (.0219) | .1005* (.0158) | 24.37 (2.20) | .4952* (.1067) | .5178* (.0949) |
| | | DANCE [11] 2025 | 28.33* (1.86) | .8483* (.0352) | .0684* (.0166) | 27.33* (0.81) | .8449* (.0153) | .0633* (.0114) | 23.89* (2.12) | .4786* (.1028) | .4787* (.1019) |
| | | DiffMSR [12] 2024 | 26.77* (1.56) | .8137* (.0370) | .1397* (.0322) | 27.02* (1.02) | .8190* (.0490) | .1295* (.0235) | 23.88* (2.11) | .4743* (.1054) | .5407* (.0892) |
| | | McMRSR [8] 2022 | 28.90* (1.69) | .8561* (.0335) | .1189* (.0313) | 27.57* (0.91) | .8406* (.0289) | .1088* (.0188) | 24.32 (2.18) | .4887* (.1065) | .5287* (.0911) |
| | | SAN [29] 2022 | 28.20* (1.66) | .8391* (.0339) | .1341* (.0308) | 27.05* (0.90) | .8239* (.0169) | .1774* (.0247) | 23.97* (2.12) | .4752* (.1059) | .5289* (.0854) |
| | SISR | SwinIR [39] 2021 | 26.26* (1.11) | .7328* (.0317) | .1984* (.0299) | 24.76* (0.87) | .8204* (.1153) | .2897* (.0169) | 23.22* (1.99) | .4501* (.1032) | .6000* (.0795) |

TABLE II
QUANTITATIVE COMPARISON OF OUR PROPOSED METHOD ("OURS") AGAINST STATE-OF-THE-ART METHODS AT THE SCALE FACTORS OF 4× AND 8×. PERFORMANCE IS EVALUATED ON THE PATHOLOGY REGIONS AND THE NON-PATHOLOGY REGIONS IN FASTMRI USING PSNR ↑, SSIM ↑, AND LPIPS ↓ BY THEIR MEAN AND (STANDARD DEVIATION). THE BEST RESULT FOR EACH METRIC IS HIGHLIGHTED IN RED. THE SECOND BEST IS HIGHLIGHTED IN BLUE.
* DENOTES STATISTICALLY SIGNIFICANT DIFFERENCE FROM OURS FOR THE SAME DATASET, SCALE FACTOR AND METRIC (P<0.01).

| | Pathology Regions | | | | | | Non-Pathology Regions | | | | | |
|---|---|---|---|---|---|---|---|---|---|---|---|---|
| | 4× | | | 8× | | | 4× | | | 8× | | |
| | PSNR↑ | SSIM↑ | LPIPS↓ | PSNR↑ | SSIM↑ | LPIPS↓ | PSNR↑ | SSIM↑ | LPIPS↓ | PSNR↑ | SSIM↑ | LPIPS↓ |
| Ours | 28.30 (3.06) | .7264 (.0956) | .3680 (.0996) | 24.96 (3.13) | .5395 (.1231) | .6192 (.0941) | 27.73 (2.19) | .6837 (.0912) | .3835 (.0702) | 24.20 (1.76) | .5432 (.0877) | .5848 (.0699) |
| A2CDic [15] 2025 | 26.96* (3.12) | .6870* (.1016) | .3729 (.1252) | 24.87 (2.93) | .5423 (.1175) | .4155* (.1299) | 25.89* (2.41) | .5993* (.0985) | .4309* (.0932) | 24.20 (2.07) | .4919* (.0981) | .5173* (.0867) |
| DANCE [11] 2025 | 26.36* (3.14) | .6614* (.1029) | .3252* (.1089) | 24.21* (2.90) | .5099* (.1197) | .3976* (.1271) | 25.34* (2.36) | .5798* (.0983) | .3782* (.0803) | 23.70* (1.99) | .4749* (.0942) | .4762* (.0935) |
| DiffMSR [12] 2024 | 26.82* (3.09) | .6748* (.1011) | .3800* (.1177) | 24.30* (2.90) | .4866* (.1177) | .4572* (.1170) | 25.85* (2.37) | .5937* (.0978) | .4359* (.0863) | 23.73* (1.98) | .4714* (.0972) | .5414* (.0818) |
| McMRSR [8] 2022 | 26.99* (3.09) | .6838* (.1019) | .3751 (.1236) | 24.80* (2.88) | .5291* (.1176) | .4274* (.1283) | 25.98* (2.41) | .5983* (.0985) | .4338* (.0908) | 24.16 (2.05) | .4853* (.0977) | .5288* (.0832) |
| SAN [29] 2022 | 26.82* (3.10) | .6728* (.1021) | .3756 (.1182) | 24.40* (2.98) | .4901* (.1188) | .4414* (.1136) | 25.86* (2.37) | .5921* (.0981) | .4361* (.0868) | 23.82* (1.99) | .4717* (.0973) | .5300* (.0780) |
| SwinIR [39] 2021 | 26.19* (2.99) | .6346* (.1014) | .4089* (.1038) | 23.49* (2.79) | .4188* (.1168) | .5546* (.0997) | 25.39* (2.22) | .5732* (.0955) | .4548* (.0774) | 23.08* (1.84) | .4461* (.0941) | .6015* (.0715) |

## A. Comparison with State-of-the-Art Methods

Quantitative and qualitative comparisons for 4× and 8× super-resolution are reported in Table I and Fig. 3, respectively. Our method consistently outperforms all competing approaches, with clear performance gains at 4× and an even larger margin at 8×.

Although some methods achieve marginally higher SSIM at 4×, this reflects a known trade-off: approaches heavily optimized for L1/SSIM tend to favor structurally smooth reconstructions. In such cases, elevated SSIM in the absence of corresponding improvements in PSNR and LPIPS often reflects over-regularization, where reconstructions increasingly conform to smooth structural priors at the expense of fidelity and perceptual detail. In contrast, our method achieves superior PSNR and LPIPS, indicating sharper reconstructions with more faithful structural details (Fig. 3). This behavior is further corroborated by our ablation analysis in Fig. 8 (Section IV.E), where increasing the prior consistency weight leads to monotonic SSIM improvement accompanied by degraded

TABLE III
QUANTITATIVE COMPARISON OF OUR PROPOSED METHOD ("OURS") AGAINST THE VANILLA VARIANTS OF ITS CONSTITUENT PARTS (CYCLEGAN AND VANILLA INR ("VANINR") AT THE SCALE FACTORS OF 4× AND 8×. PERFORMANCE IS EVALUATED ON IXI, HCP AND FASTMRI USING PSNR↑, SSIM↑, AND LPIPS↓ BY THEIR MEAN AND (STANDARD DEVIATION). FOR ALL NON-OURS ENTRIES, THE DIFFERENCE FROM OURS IS STATISTICALLY SIGNIFICANT FOR THE SAME DATASET, SCALE FACTOR AND METRIC (P<0.01)

| | | IXI | | | HCP | | | FastMRI | | |
|---|---|---|---|---|---|---|---|---|---|---|
| | | PSNR↑ | SSIM↑ | LPIPS↓ | PSNR↑ | SSIM↑ | LPIPS↓ | PSNR↑ | SSIM↑ | LPIPS↓ |
| | CycleGAN | 30.72 (3.41) | **.9418 (.0205)** | .0165 (.0076) | 27.57 (0.81) | .8204 (.1153) | **.0194 (.0037)** | 18.96 (2.40) | .3745 (.0961) | **.1824 (.0596)** |
| 4× | Ours | **35.16 (2.08)** | .9319 (.0133) | **.0123 (.0052)** | **30.75 (0.88)** | **.9070 (.0151)** | .0232 (.0066) | **27.97 (2.21)** | **.6708 (.0997)** | .3841 (.0703) |
| | vanINR | 28.99 (1.45) | .8121 (.0294) | .1525 (.0304) | 27.52 (0.51) | .8185 (.0138) | .1930 (.0146) | 24.10 (2.16) | .4639(.1110) | .5252(.0889) |
| 8× | Ours | **32.53 (2.47)** | .8842 (.0194) | **.0160 (.0068)** | **28.13 (0.64)** | **.8576 (.0204)** | .0196 (.0036) | **24.35 (1.77)** | **.5493 (.0991)** | .5773 (.0707) |
| | vanINR | 25.22 (1.39) | .6067 (.0476) | .2850 (.0381) | 23.60 (0.59) | .6364 (.0228) | .3909 (.0194) | 23.13 (1.85) | .4243 (.1065) | .6421 (.0868) |

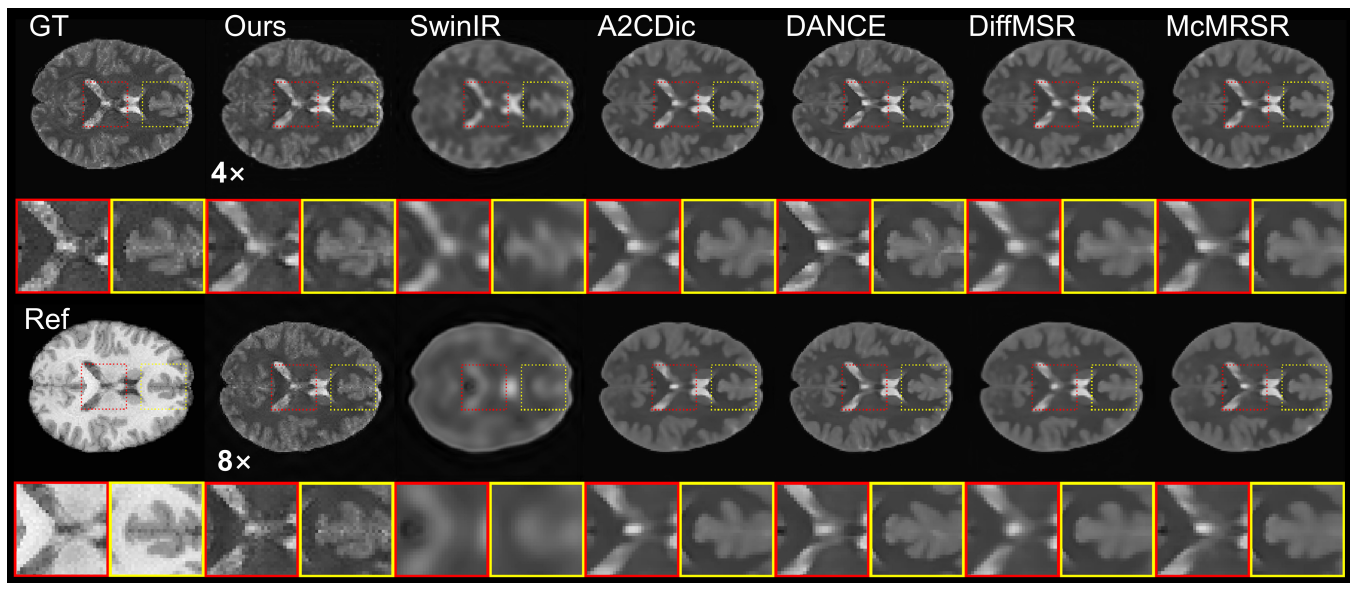

Fig. 3. Qualitative Comparison of Results of our proposed method ("Ours") on HCP dataset against SOTA methods at the scale factors of 4× and 8×.

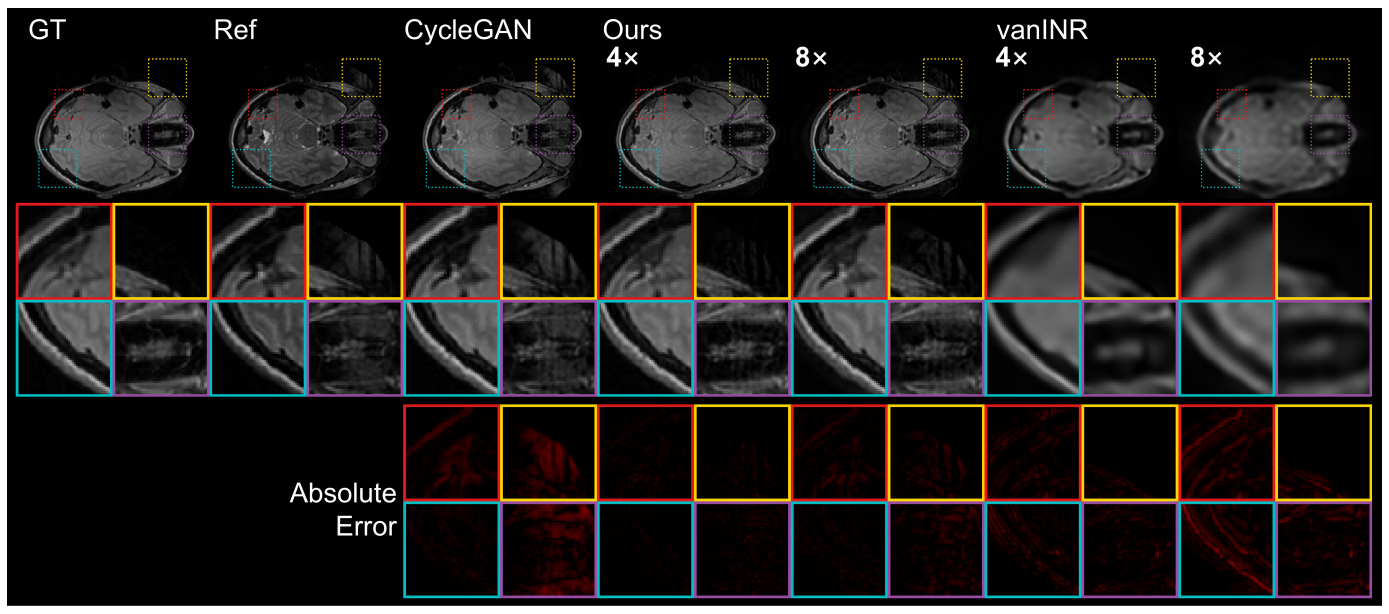

Fig. 4. Qualitative comparison on the IXI dataset of the proposed MCSR method ("Ours") with its vanilla counterparts, including CMS (CycleGAN) and unsupervised SISR (vanINR). CMS exhibits residual artefacts inherited from the reference modality (Ref) in the background (**top right yellow box**) and in the imaged organ (**bottom right purple box**), which are effectively suppressed by Ours via the IrR module. **Top left red box**: MCSR restores an artificially attenuated dark gap region in CMS results inherited from Ref. **Bottom left blue-green box**: MCSR preserves fine, accurate structural details at skull boundaries from Ref via CMS, which are not recovered by SISR vanINR particularly at 8×.

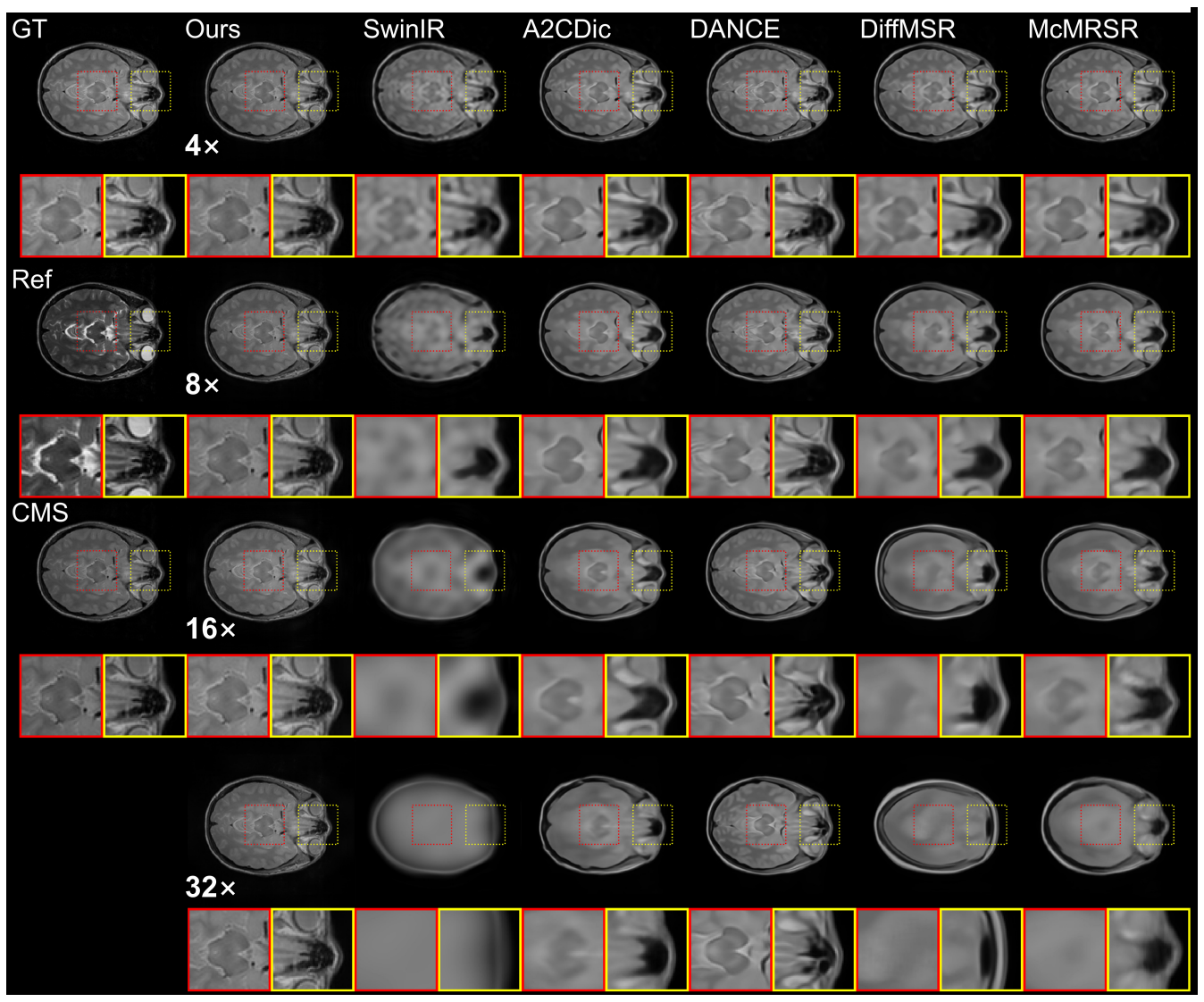

Fig. 5. Qualitative comparison on the IXI dataset. Results of the proposed method (Ours) are compared with state-of-the-art MCSR and SISR methods at scale factors of 4×, 8×, 16×, and 32×, and with the CMS constituent component of our framework. As the scale factor increases, existing SOTA MCSR methods exhibit severe degradation, including hallucinations and structural failures, particularly in the periventricular region (**red box**) and the anterior skull base (sinonasal) region (**yellow box**). In contrast, our method produces stable, anatomically consistent reconstructions across all scales, with CMS serving as a robustness guarantee under extreme upsampling conditions.

PSNR and LPIPS, a characteristic signature of over-regularization.

We further report quantitative metrics for the pathology regions and non-pathology regions in the FastMRI [36] dataset (Table II), as defined by the FastMRI+ [37] clinical pathology annotations. The pathology-region results show trends broadly consistent with the whole-image FastMRI evaluation in Table I and non-pathology region results in Table II.

## B. Comparison With Vanilla Constituent Part Variants

Table III compares our proposed full framework with its vanilla constituent variants, including the unpaired CycleGAN (uCMS) and the vanilla INR (vanINR). Our method consistently achieves the highest PSNR across datasets and scale factors, demonstrating superior reconstruction fidelity. Although CycleGAN occasionally attains higher SSIM or LPIPS, such improvements accompanied by reduced PSNR often indicate oversmoothing or hallucinated structures rather than faithful recovery.

A key concern in staged MCSR frameworks is that the unpaired nature of uCMS may introduce residual artefacts or modality-inconsistent signals inherited from the reference modality, which can propagate into the synthesized prior and potentially bias the final reconstruction. As shown in Fig. 4, the IrR module effectively suppresses these residual artefacts in both the background (Fig. 4: top right yellow box) and the imaged organ (Fig. 4: bottom right purple box) by enforcing consistency with the subject's acquired LR target. Moreover, IrR corrects the artificially attenuated dark-gap artefact (Fig. 4: top left red box) that arises from modality-dependent contrast differences, where tissues exhibit suppressed signal in the reference modality. The bottom left blue-green box (Fig. 4) further illustrates that IrR successfully preserves and refines fine anatomical details at tissue boundaries, which are not recovered by the SISR-only vanINR, particularly at 8×. At the same time, Fig. 4 also indicates that this correction is not unlimited: at higher magnification factors such as 8×, where the LR target provides weaker subject-specific constraints, residual prior-driven errors are less fully suppressed than at 4×. This behavior supports our interpretation of uCMS as a soft prior and IrR as a data-consistent correction mechanism, while also highlighting that stronger synthesis bias becomes more difficult to regularize under severe information scarcity.

These results highlight the strong synergy of our decoupled design: uCMS provides a robust anatomical prior, while IrR enforces patient-specific consistency and suppresses artefacts, yielding reconstructions that substantially outperform either constituent component alone.

### C. Robustness to Extreme Scale Factors

A key claim of our work is the ability to perform robust, scale-agnostic super-resolution. We validate this by evaluating all methods across a challenging range of scale factors (4×, 8×, 16×, and 32×), with results shown in Fig. 5 and Fig. 6.

As the scale factor increases, the performance of all methods degrades; however, our framework exhibits substantially greater robustness across all metrics. Fig. 5 clearly shows that at higher scales (particularly at 8× and above), SISR methods suffer from severe feature loss, while most MCSR methods become strongly influenced by the extremely sparse LR target, leading to structural collapse or hallucinations. These failures are most evident in anatomically sensitive regions such as the periventricular area (Fig. 5: red box) and the anterior skull base (sinonasal) region (Fig. 5: yellow box). In contrast, our method maintains stable, anatomically consistent reconstructions across all scales, with CMS acting as a robustness guarantee that prevents degradation under extreme upsampling, even when the LR input provides minimal guidance.

The quantitative trends in Fig. 6 further support this observation. For PSNR and SSIM, all competing methods exhibit a steep, near-linear decline as the scale increases, whereas our method shows a much flatter degradation curve, maintaining PSNR above 32 dB even at 32× – a regime in which most other methods effectively fail (PSNR < 24 dB).

The LPIPS plot is particularly revealing. Both our method and the CycleGAN baseline maintain consistently low (better) LPIPS scores across all scales, indicating that the uCMS module provides a stable, perceptually meaningful prior that is largely independent of the target resolution. In contrast, methods relying purely on self-supervision, such as vanilla INR, suffer a catastrophic loss of perceptual quality as the scale factor increases and the LR signal becomes insufficient, while existing MCSR methods also exhibit pronounced degradation due to their sensitivity to extreme LR sparsity.

Together, these results confirm our hypothesis that combining a strong generative population prior with a scale-agnostic implicit representation is essential for achieving robust and reliable super-resolution at extreme magnification factors.

While super-resolution factors such as 16× and 32× exceed typical clinical acquisition requirements, we emphasize that these experiments are intended as *controlled stress tests* rather than direct clinical targets. In practice, clinically relevant super-resolution factors are typically much lower (e.g., 2×–8×). Extreme upsampling magnifies the ill-posedness of the reconstruction problem and provides a sensitive probe of how different methods balance prior information against data fidelity. In this regime, failure modes such as structural collapse, hallucination, or excessive smoothing become particularly evident. The strong performance of the proposed method under these conditions highlights the robustness of the decoupled prior-reconstruction formulation.

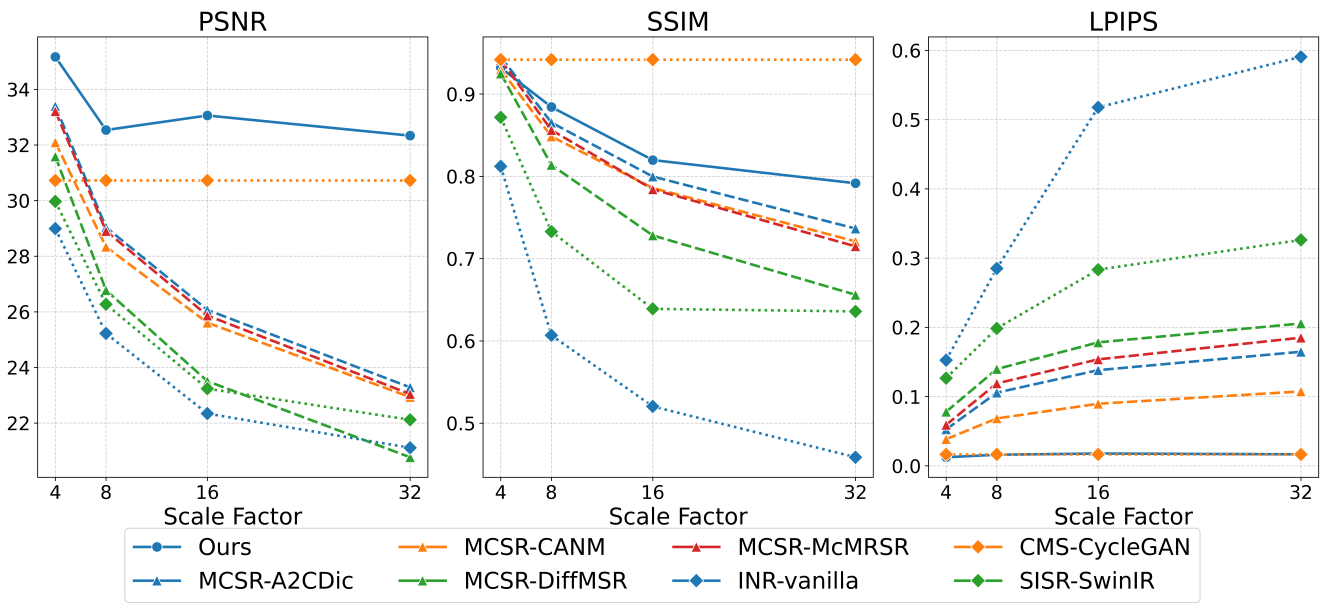

**Fig. 6.** Quantitative comparison of super-resolution performance across extreme scale factors (4×, 8×, 16×, 32×) on the IXI dataset. The plots for PSNR (left), SSIM (middle), and LPIPS (right) show that while all competing methods (dashed lines) degrade sharply, our framework ("Ours", solid blue) is substantially more robust. It maintains the highest PSNR and SSIM and the lowest (best) LPIPS, confirming its effectiveness at high magnifications.

### D. Computational Efficiency

Beyond accuracy, our framework is computationally efficient per optimization step. Fig. 7 compares reconstruction fidelity (PSNR) against computational cost (GFLOPs, log-scale) and model size (number of parameters) on the IXI dataset. Because INR requires iterative test-time fitting, the GFLOPs reported in Fig. 7 correspond to the per-iteration forward-pass cost; we therefore additionally report wall-clock optimization time to characterize end-to-end runtime. Our method lies in the desirable top-left region, achieving the highest PSNR while using one of the smallest models and the lowest per-forward-pass computational cost. Specifically, the per-subject IrR module uses 0.72M parameters and 94.5 GFLOPs per iteration, which is substantially smaller than MCSR baselines such as A2CDic (9.9M parameters, 2.14 TFLOPs) and McMRSR (2.1M parameters, 563.8 GFLOPs).

This efficiency stems from our decoupled design: the large uCMS prior (e.g., CycleGAN) is trained once offline, while patient-specific adaptation is performed exclusively using the lightweight IrR network. Importantly, IrR has the same capacity and per-step computational cost as the INR(vanilla) baseline yet achieves a +6.17 dB PSNR improvement (35.16 vs. 28.99) with no increase in model size or per-iteration compute, due to the superior initialization provided by the uCMS prior.

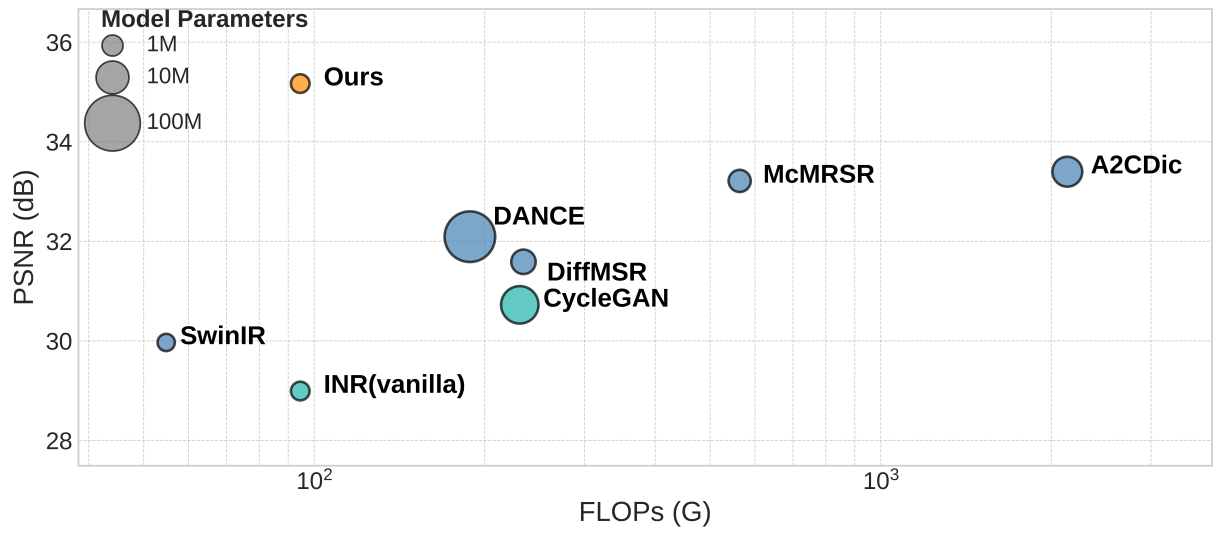

**Fig. 7.** Model performance vs. computational efficiency for the 4× SR task on the IXI dataset. The y-axis represents reconstruction fidelity (PSNR), the x-axis represents computational cost (GFLOPs, log-scale), and the bubble size represents the number of model parameters. Our method occupies the ideal top-left quadrant, demonstrating state-of-the-art performance with a model that is orders of magnitude smaller and more efficient than competing SOTA MCSR approaches. Detailed quantitative performance values for the corresponding methods can be found in Table I.

Like other patient-specific INR-based approaches, the proposed framework requires test-time optimization, resulting in longer per-subject inference time compared to feed-forward models. Under the default configuration, optimization takes approximately 55.7 seconds per 2D slice on an RTX 3090 GPU, which is comparable to existing self-supervised INR methods. This cost is partially offset by the fact that the uCMS prior is trained once offline and reused across subjects, and that no paired-data training is required. Future work will explore acceleration strategies such as warm-start initialization, multi-resolution optimization, and parallelized slice or patch-based fitting to further reduce inference time.

### E. Ablation Study: Influence of Prior Weighting

To analyze the interplay between loss terms in our objective function in (4), we conduct an ablation study on the IXI dataset by varying the prior consistency weight $\alpha$. The data fidelity weight $\beta$ is fixed to 1.0, while $\alpha$ is swept over several orders of magnitude. Quantitative results for PSNR, SSIM, and LPIPS at all four upscaling factors are shown in Fig. 8.

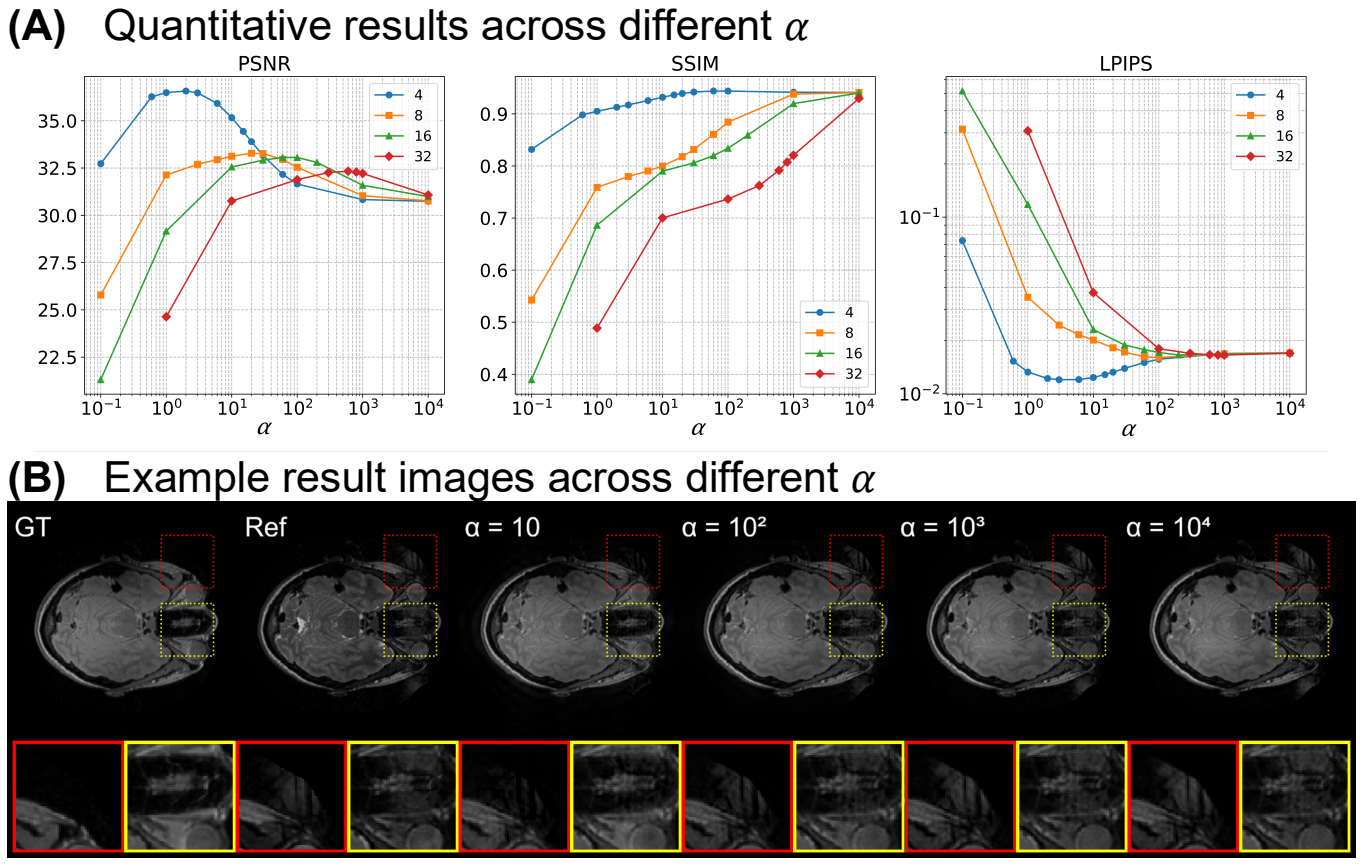

**Fig. 8.** Ablation study on the influence of the prior consistency weight $\alpha$ as in (4) on the IXI dataset. (A) Quantitative results are reported in terms of PSNR, SSIM, and LPIPS for all four upscaling factors. The x-axis (log scale) denotes the value of $\alpha$, while the data fidelity weight $\beta$ is fixed to 1.0. PSNR and LPIPS exhibit a clear optimal operating region, beyond which performance degrades due to over-regularization by the prior. In contrast, SSIM increases monotonically with $\alpha$, reflecting improved structural similarity but reduced sensitivity to fidelity loss at large $\alpha$. The selected scale-specific $\alpha$ values lie near the joint optimal region of PSNR and LPIPS while achieving high SSIM, providing a balanced trade-off across all three metrics. (B) Qualitative visual examples of result images across different orders of $\alpha$.

Distinct and metric-dependent trends are observed. Both PSNR and LPIPS exhibit a clear optimal operating region as $\alpha$ increases. PSNR shows a pronounced unimodal peak, while LPIPS reaches a minimum, although the latter is comparatively flatter and becomes more apparent when evaluated on a logarithmic scale. Beyond this region, further increasing $\alpha$leads to degradation in both PSNR and LPIPS, indicating a loss of fidelity due to excessive reliance on the synthetic prior.

In contrast, SSIM increases monotonically (or saturates) with larger $\alpha$, reflecting its sensitivity to structural similarity. While higher SSIM generally corresponds to improved structural coherence, the divergence between SSIM and the fidelity-sensitive metrics (PSNR and LPIPS) at large $\alpha$ suggests that SSIM alone may not fully capture reconstruction fidelity in this regime. In particular, continued SSIM improvement accompanied by declining PSNR and LPIPS is indicative of over-regularization, where reconstructions increasingly conform to the prior $\widetilde{I_{hr}}$ rather than the patient-specific LR measurements $I_{\mathrm{lr}}$ . This behavior may correspond to the introduction of overly smooth or prior-driven structures.

At the other extreme, when $\alpha$ is too small (e.g., $10^{-3}$), the optimization is dominated by the data fidelity term $\mathcal{L}_{\mathrm{data}}$ , effectively reducing the model to an INR-vanilla formulation. In this case, the sparse LR observations are insufficient to recover high-frequency anatomical details, particularly at higher upscaling factors, resulting in degraded performance across all metrics.

Based on these observations, we select $\alpha$ values that lie near the joint optimal region of PSNR and LPIPS, while favoring higher SSIM within this region. This choice reflects a principled compromise between fidelity, perceptual quality, and structural consistency. The resulting scale-specific $\alpha$ values for the IXI dataset (10.0 (4×), 100.0 (8×), 60.0 (16×), and 600.0 (32×)) are validated by Fig. 8, where they consistently fall within the balanced operating region across all three metrics. In practice, $\alpha$ is selected once for each upsampling factor on a held-out validation setting and then fixed across subjects, rather than tuned separately for each patient. Importantly, because Fig. 8 is shown on a logarithmic $\alpha$-axis, the effective near-optimal region is relatively broad, indicating that moderate deviations from the selected value do not cause abrupt performance deterioration.

## V. Discussion

This work introduces a new perspective on multi-contrast super-resolution by explicitly decoupling population-level cross-modal prior learning from patient-specific resolution recovery. Unlike conventional MCSR approaches that entangle cross-modal feature extraction and super-resolution within a single supervised model, we reformulate MCSR as the combination of CMS and target-domain re-representation, enabling the two components to be trained independently and under fundamentally different data assumptions.

The proposed framework still operates in the standard MCSR setting, where a subject-matched HR reference contrast and LR target contrast are jointly available at inference. The practical advantage lies in training: unlike conventional supervised MCSR approaches, the method does not require paired population-level target LR/HR data or paired cross-modal HR reference-target data to learn the population prior. Accordingly, 'unpaired' in this work should be understood as referring to the training-data assumption of the framework rather than to fully unpaired subject-specific reconstruction.

The proposed framework effectively bridges two previously disjoint paradigms. Unpaired CMS leverages large, heterogeneous population datasets to learn a robust anatomical prior without requiring cross-modal correspondence, while the IrR module enforces strict subject-specific data fidelity through self-supervision. Importantly, the CMS output is not treated as a final reconstruction, nor is it assumed to be anatomically reliable in isolation. As an unpaired generative model, uCMS may introduce biased contrast mappings or hallucinated structures inherited from population statistics or reference-modality artefacts. In the proposed framework, the CMS output is therefore used exclusively as a *soft population-level prior*, which is explicitly corrected during the implicit re-representation stage through an analytic data fidelity constraint. This design ensures that synthesized features inconsistent with the subject's acquired LR measurements are actively suppressed, rather than propagated into the final reconstruction. Nevertheless, this correction mechanism is fundamentally limited by the information content of the observed LR target; when the prior mismatch is severe and the target observation is highly sparse, residual prior-driven errors may still persist.

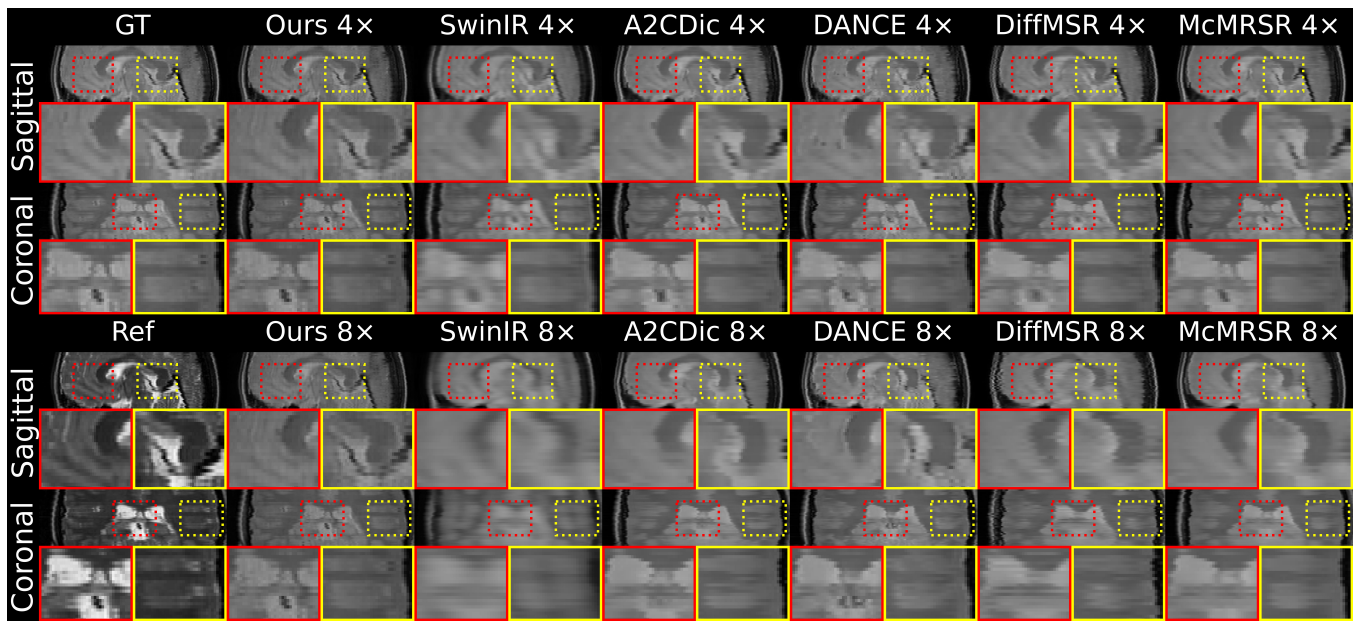

Fig. 9. Qualitative comparison of central sagittal and coronal reformations of the axially reconstructed volumes of our proposed method ("Ours") on IXI dataset against SOTA methods at the scale factors of 4× and 8×.

A central advantage of this decoupled formulation is its robustness under extreme information scarcity within the controlled retrospective degradation setting evaluated in this study. At high upsampling factors (16× and 32×), existing MCSR methods become increasingly unstable due to their reliance on sparse LR targets or fixed-scale supervised priors. In contrast, our results demonstrate that combining a strong generative population prior with a scale-agnostic implicit representation yields stable and anatomically consistent reconstructions even when the LR signal provides minimal guidance. While such extreme magnification factors may exceed routine clinical requirements, they serve as a stringent stress test that reveals fundamental differences in robustness and prior utilization across methods. Accordingly, the robustness demonstrated here should be interpreted with respect to severe resolution loss under the adopted retrospective k-space truncation model, rather than as evidence of robustness to the full variability of prospective clinical LR acquisitions.

The ablation analysis further highlights the importance of balancing population-level priors and subject-specific fidelity. Excessive reliance on the CMS prior leads to over-regularization, while insufficient weighting reduces the framework to a vanilla INR formulation that struggles under severe undersampling. These observations emphasize that effective MCSR is not achieved by stronger priors alone, but by carefully controlled interaction between prior knowledge and measured data. Although the prior-consistency weight $\alpha$ is scale-dependent, our ablation indicates a broad near-optimal operating region on a logarithmic scale, suggesting that fixed scale-specific defaults are sufficient in the current setting; fully automatic subject-specific selection remains future work.

Despite its advantages, this work has several limitations. First, all experiments are conducted on retrospective 2D slice-based reconstructions with simulated k-space truncation. We adopt this setting because it is the standard controlled protocol in current MRI super-resolution and multi-contrast super-resolution benchmarking, enabling direct comparison with prior methods. At the same time, this setting may not fully capture the full continuity across the $z$-axis (through-plane) or the variability of prospective clinical LR acquisitions, including scanner- and sequence-specific resolution characteristics, acquisition noise, motion corruption, partial-volume effects, and other acquisition-dependent artefacts. Accordingly, the robustness demonstrated in this study should be interpreted within the controlled retrospective k-space truncation setting and does not establish robustness to the full variability of real-world clinical LR acquisitions. Validation under more realistic degradation models and prospectively acquired LR data remains an important direction for future work. Since clinical MRI interpretation is usually volumetric, the commonly used slice-wise reconstruction may also introduce potential inter-slice inconsistencies. To provide an additional qualitative assessment, we here provide sagittal and coronal reformations of the axially reconstructed volumes (Fig. 9). Second, the effectiveness of the population prior depends on the representativeness of the unpaired training dataset; extreme domain shifts, extreme spatial misregistration between reference and target contrasts, or rare anatomical variations may reduce the reliability of the synthesized prior. Third, the requirement for test-time optimization introduces additional computational overhead relative to feed-forward models. Finally, we emphasize that the present evaluation is methodological rather than reader-study-based. Accordingly, PSNR, SSIM, LPIPS, and qualitative visual comparisons are used here to assess reconstruction fidelity and appearance under controlled benchmark conditions, where future work will explore the results with review from radiologist experts. These limitations motivate future extensions toward 3D representations, diverse unpaired training corpora, and faster optimization strategies. Future work will explore acceleration strategies such as warm-start initialization from anatomically similar subjects or from a pre-trained implicit prior, as well as multi-resolution fitting and further parallelization.

## VI. Conclusion

We presented a decoupled, data-efficient framework for multi-contrast MRI super-resolution that unifies unpaired population-level prior learning with patient-specific implicit re-representation. By reformulating MCSR as a composition of CMS and target-domain re-representation, the proposed method eliminates the need for paired population-level target LR/HR supervision and paired cross-modal HR reference-target supervision while achieving superior reconstruction fidelity and exceptional robustness at extreme upsampling factors. This work establishes a flexible paradigm for MCSR and highlights the importance of explicitly separating prior learning from data-consistent reconstruction in medical image super-resolution.